\documentclass[letterpaper, 10 pt, conference]{ieeeconf}  

\IEEEoverridecommandlockouts                              

\usepackage{times}
\usepackage{epsfig}
\usepackage{graphicx}
\usepackage{amsmath}
\usepackage{amssymb}
\usepackage{footnote}
\makesavenoteenv{tabular}

\title{\LARGE \bf
Goal-oriented Object Importance Estimation in On-road Driving Videos
}

\author{Mingfei Gao$^{1*}$, Ashish Tawari$^{2}$ and Sujitha Martin$^{2}$
\thanks{$^{*}$Work done during an internship at the Honda Research Institute, USA.}
\thanks{$^{1}$The author is with the University of Maryland, College Park, MD, 20740. {\tt\small mgao@umiacs.umd.edu}}%
\thanks{$^{2}$The authors are with the Honda Research Institute, Mountain View, CA, 94043.
        {\tt\small \{atawari, smartin\}@honda-ri.com}}%
}

\begin{document}

\maketitle
\thispagestyle{empty}
\pagestyle{empty}

\begin{abstract}
We formulate a new problem as Object Importance Estimation (OIE) in on-road driving videos, where the road users are considered as important objects if they have influence on the control decision of the ego-vehicle's driver. The importance of a road user depends on both its visual dynamics, \emph{e.g}., appearance, motion and location, in the driving scene and the driving goal, \emph{e.g}., the planned path, of the ego vehicle. We propose a novel framework that incorporates both visual model and goal representation to conduct OIE. To evaluate our framework, we collect an on-road driving dataset at traffic intersections in the real world and conduct human-labeled annotation of the important objects. Experimental results show that our goal-oriented method outperforms baselines and has much more improvement on the left-turn and right-turn scenarios. Furthermore, we explore the possibility of using object importance for driving control prediction and demonstrate that binary brake prediction can be improved with the information of object importance.
\end{abstract}

\section{Introduction}
\label{sec: intro}
Human's vision system plays a key role for perceiving and interacting with traffic participants under the complicated driving context. When looking into the dynamic scene, a driver can rapidly select the objects that are relevant for the driving task and make a control decision for effective and efficient driving. Inspired by this visual selection mechanism, driver's attention has been studied in recent years in order to understand the human driving behavior and ultimately help the driving control system of autonomous vehicles. Existing works focus on pixel-level driver's attention prediction by mimicking human gaze behavior~\cite{dreyeve2018,tawari2017computational,xia2017training}. However, there are at least two drawbacks of using human gaze: 1) human gaze is sometimes not directly related to the driving task. For example, drivers may look at the billboards for their own interests; 2) human gaze is sequential which makes it impossible to capture all the important information at the same time. Moreover, existing works only take the perceived driving video as input and do not consider the effect of the driver's goal, while driver's goal is an essential factor to select relevant objects. For example, objects relevant for making control decisions  should be very different when the ego vehicle is turning right versus turning left.

\begin{figure}[t]
\begin{center}
   \includegraphics[width=0.9\linewidth]{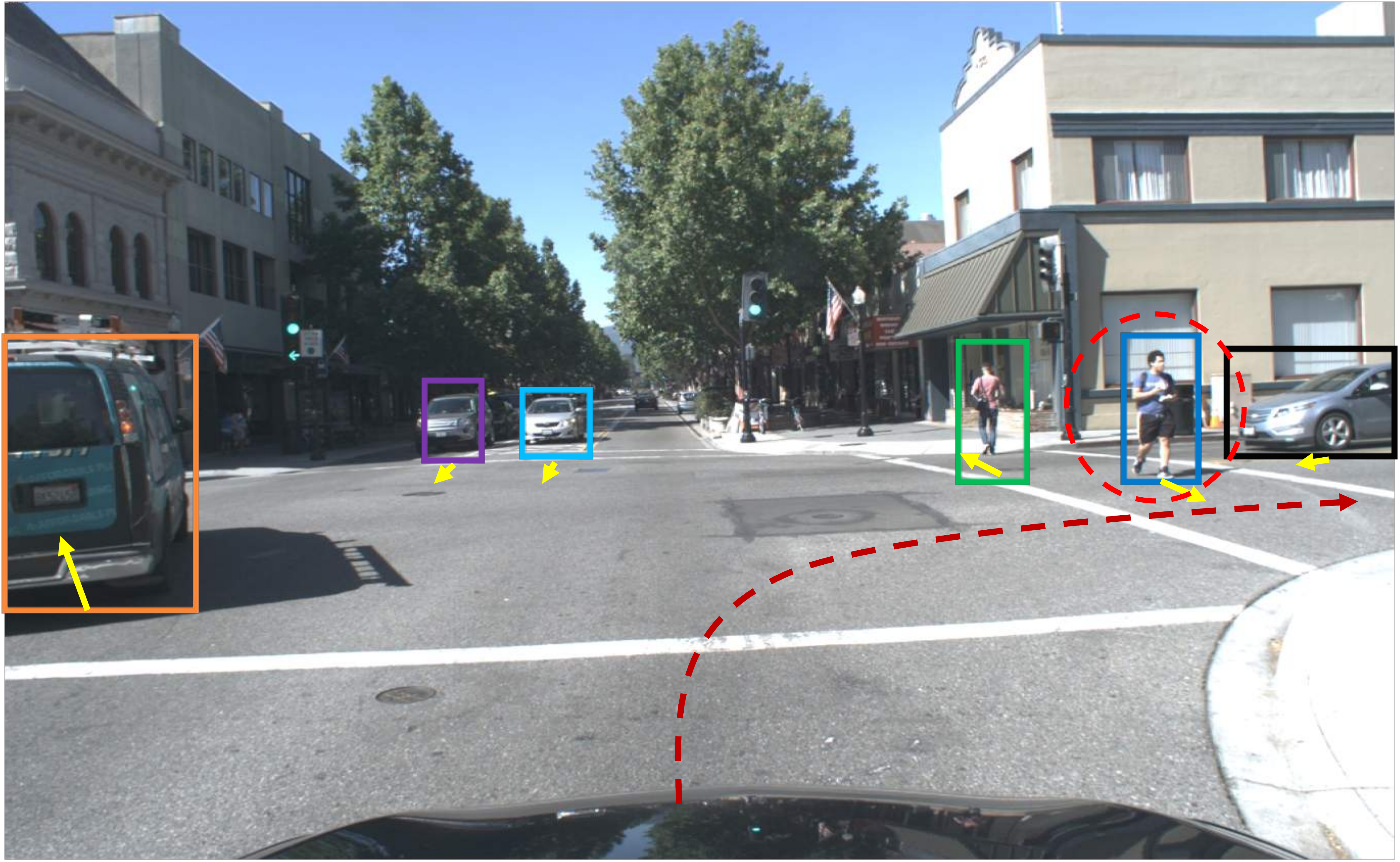}
\end{center}
   \caption{The scenario of our work. Bounding boxes with arrows indicate the moving road users, dotted line shows the planned path of the ego vehicle and the dotted circle includes the important object. Given the dynamic status of the road users, a driver's driving-related attention usually lands on the road users that have influence on the control decision of the driver. Moreover, the attention highly depends on the driving goal of the vehicle.}
\label{fig: idea}
\end{figure}

To handle those limitations, we formulate the problem as Object Importance Estimation (OIE) in on-road driving videos. The important objects are defined as the road users, \emph{i.e.}, vehicles and persons, that are relevant for the ego vehicle's driver to make the vehicle control decision. Our definition ensures that the important objects are directly related to the driving task and that multiple important objects can be captured at the same time. Static semantic driving context, \emph{e.g.}, traffic lights, line marks and drivable areas, can also influence the driving behavior. However, we only focus on the interactions with the road users and leave the static semantic driving context for future work. Fig.~\ref{fig: idea} shows an example of the scenario that our work focuses on. Visual dynamics of road users are important for our model to understand the driving scene. Also, the driver's goal (where the vehicle is going) is essential for object importance estimation. For example, in Fig.~\ref{fig: idea}, if the ego vehicle is turning left instead, all the pedestrians on the cross walk at the right side will not be as important to the ego vehicle.

To solve the proposed OIE problem, we present a novel framework where both the features of the dynamic road users (\emph{visual model}) and the driving goal (\emph{goal model}) are incorporated. In order to evaluate our framework, we collect an on-road driving dataset in the real world and annotate the important objects given the context. To provide more complex interactions between the road users and the ego vehicle, our dataset focuses on traffic intersections. Experiments show that our method largely outperforms the baselines, especially for the scenarios that the ego vehicle is turning left/right which demonstrates that modeling the driving goal is very important for our task. To explore the possibility of using important objects to improve driving control prediction, we conduct an experiment on binary brake prediction. Results show that the binary brake prediction can be improved with the information of the object importance.

\begin{figure}[t]
\begin{center}
   \includegraphics[width=1.0\linewidth]{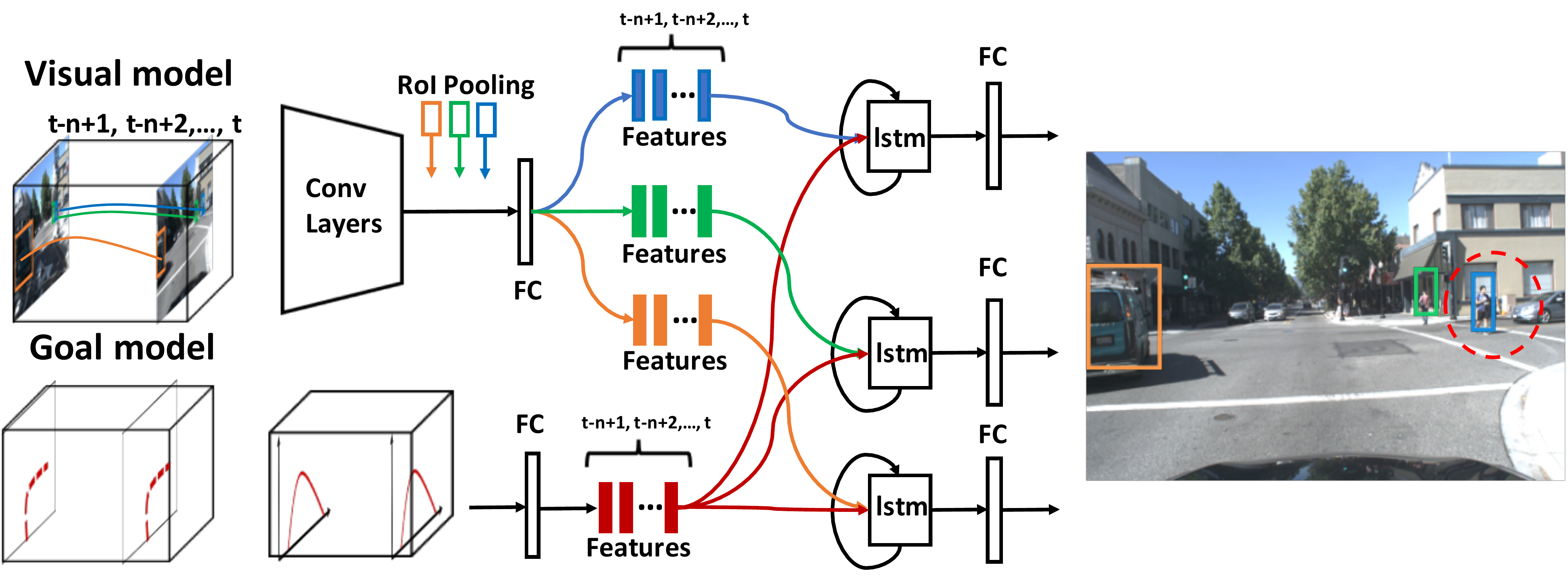}
\end{center}
   \caption{The proposed approach has two branches, \emph{e.g.}, visual model and goal model. Object tracking is done for all the road users through the input clip. Visual features of objects are extracted at each time step. Goal model describes the driving goal at each time step using sampled points on the planned path in the real world. A common goal-oriented feature is concatenated with features of each object at the corresponding time to form the final feature representation. A shared LSTM model is used to predict the importance score for every object given the final features. Objects and their features are differentiated using different colors.}
\label{fig: pipeline}
\end{figure}
\section{Related Works}
\subsection{Driver's Attention Prediction}
\textbf{Human Gaze based Approach}. Existing works focus on driver's attention prediction supervised by human gaze information~\cite{tawari2017computational,dreyeve2018,xia2017training}. Tawari and Kang propose a Bayesian framework for driver's attention prediction where a fully convolutional network is utilized with only images as input in~\cite{tawari2017computational}. Palazzi \emph{et al}. proposed a multi-branch model that incorporates RGB, optical flow and semantic segmentation clips in~\cite{dreyeve2018} and C3D~\cite{tran2015learning} is used to extract features from multiple branches. In~\cite{xia2017training}, Xia \emph{et al}. propose a driver's attention framework where a human weighted sampling strategy is used during training to handle critical situations. Kim \emph{et al}. explore the idea of using driver's attention to interpret the driving control prediction in~\cite{kim2017interpretable}.

\textbf{Driver's Attention Prediction Dataset}.
There are several datasets~\cite{simon2009alerting,underwood2011decisions,fridman2016driver,pugeault2015much,alletto2016dr} can be used for driver's attention prediction, but most of them are either restricted to limited settings or not publicly available. To the best of our knowledge, \emph{Dr(eye)ve}~\cite{alletto2016dr} is the only public on-road driving dataset for the driver's attention prediction task. It consists of 555,000 frames divided into 74 video sequences. Human gaze is captured by eye tracking glasses and projected to the corresponding on-road driving video frame. However, it is not suitable for our task, since 1) it has only per pixel saliency annotations based on human gaze which cannot be easily converted for important object labels; 2) it contains mostly scenarios of driving on the straight road (mostly the vehicle is trying to keep itself between lines or following another vehicle) which makes it not complicated enough for our task. Driving at the traffic intersections is a more appropriate scene for us, since it provides more opportunities for the ego vehicle to interact with other road users.

\subsection{Region based Object Detector}
CNN detectors have achieved great success~\cite{girshick14CVPR,girshick2015fast,ren2015faster,he2017mask,gao2018dynamic,singh2018analysis,gao2018c,zhou2018learning}. Region based CNN (R-CNN) is one of the most popular frameworks. Girshick~\emph{et al}. initially proposed the two-stage R-CNN framework in~\cite{girshick14CVPR} where object proposals are obtained first and then classified to different categories. Later, Fast R-CNN is proposed in \cite{girshick2015fast} to speed up R-CNN~\cite{girshick14CVPR} via end-to-end training/testing. However, it relies on external object proposal algorithms. Ren~\emph{et al}. present Faster R-CNN~\cite{ren2015faster} which jointly trains the proposal generation and the detection branches in a single framework. Further more, He~\emph{et al}. extend Faster R-CNN in~\cite{he2017mask} and create an unified architecture for joint detection and instance segmentation. Our problem is related to R-CNN in a sense that we also assign some scores to the proposed object candidates. However, we estimate object importance under the driving context rather than differentiating object categories, \emph{e.g.}, dog and cat.

\section{Problem Formulation}
The problem is formulated as goal-oriented object importance estimation where the inputs are on-road driving video clip and the goal of the ego vehicle. The outputs are the detected objects with importance scores at the last frame of the video clip. The planned path information which can be obtained from autonomous driving (AD) path planning module when the vehicle is driving online, is used to represent the goal of the vehicle.

Inspired by the R-CNN frameworks, we propose a two-stage framework which firstly generates object tracklinks from videos as object proposals and then classify the proposals to the binary classes, \emph{e.g.}, \emph{important object} and \emph{background}. Different from R-CNN detectors which generate proposals from static images, we track every object from the input video clip and treat the entire track link of an object as a proposal, since unlike the general object detection scenario where object categories, \emph{e.g.}, dog and cat, can be determined just from a static image, the object importance depends on the dynamics of objects through the video.

\section{Model Description}
As we mentioned in Sec.~\ref{sec: intro}, object importance depends on both the dynamic of the object itself and the driving goal of the ego vehicle. Thus, our method fuses the information from both parts. Due to the good performance of recurrent networks~\cite{xu2018temporal, yao2018egocentric,gao2019startnet} on online action detection tasks, our framework is based on LSTM~\cite{hochreiter1997long}.

Our framework is shown in Fig.~\ref{fig: pipeline}. The first branch describes our visual model. Multiple object tracking is performed on the input video clip. Thus, for each object candidate, $i$, its bounding-box location, $B^t_i$, is obtained at each time step $t$. Note that each time step corresponds to each image frame in the input video clip. For each object candidate at every time step, high dimensional features $\textbf{f}^t_i$ are extracted to represent the appearance, motion and location of the object. We use a feature matrix $\textbf{F}^t_i = [\textbf{f}^{t-n+1}_i, \textbf{f}^{t-n+2}_i,...,\textbf{f}^t_i]$ to represent each object $i$, in the video where $n$ is the length of the input clip. Without goal information, LSTM can be used directly with the $\textbf{F}^t_i$ as the input and the output is score $s^t_i$ of being an important object at time $t$. We will use it as a baseline in our experiment section.

The second branch shows our goal model. We extract the goal-oriented feature $\textbf{g}^t$ at time $t$ from the AD path planning module. The extracted feature is concatenated with the features of each object in the image to form the final feature representation $\textbf{gof}^t_i=[\textbf{f}^t_i, \textbf{g}^t]$, for the object. The representation for the object within the whole clip is $\textbf{GoF}^t_i=[\textbf{gof}^{t-n+1}_i,\textbf{gof}^{t-n+2}_i,...,\textbf{gof}^t_i]$. A one-layer LSTM model followed by a fully connected (FC) layer performs over $\textbf{GoF}^t_i$ to output the importance score for each object $i$ as shown in Eq.~\ref{eq: lstm}, where $\textbf{W}$ and $\textbf{b}$ indicate parameters of the FC layer. Softmax layer is used then to output the corresponding important probability.

\begin{equation}
\label{eq: lstm}
    \textbf{s}_i^t = \textbf{W}(LSTM(\textbf{GoF}_i^t))+\textbf{b}.
\end{equation}

\textbf{Visual Feature}. Appearance, motion and location features are combined to represent the dynamic changes of an object. Appearance feature is extracted from the \emph{fc7} layer of Faster R-CNN~\cite{ren2015faster} pretrained on the Pascal VOC2007~\cite{pascal-voc-2007} and VOC2012~\cite{pascal-voc-2012} \emph{trainval} sets with Resnet101~\cite{He2015} as the backbone. The appearance feature describes both the appearance of the object and the local context around the object~\cite{ren2015faster}. Histogram of flow~\cite{dalal2006human} with BIN=12 of each object bounding box is extracted as the motion feature. Location feature is represented by $(\frac{x^t_i}{W^t}, \frac{y^t_i}{H^t}, \frac{w^t_i}{W^t}, \frac{h^t_i}{H^t})$ where $x^t_i$, $y^t_i$, $ w^t_i$ and $h^t_i$ indicate the left-top corner of $B^t_i$, its width and height. $W^t$ and $H^t$ indicate the width and height of image $t$. The visual feature, $\textbf{f}^t_i$, is the concatenation of these three features.

\begin{figure}
\begin{center}
   \includegraphics[width=0.9\linewidth]{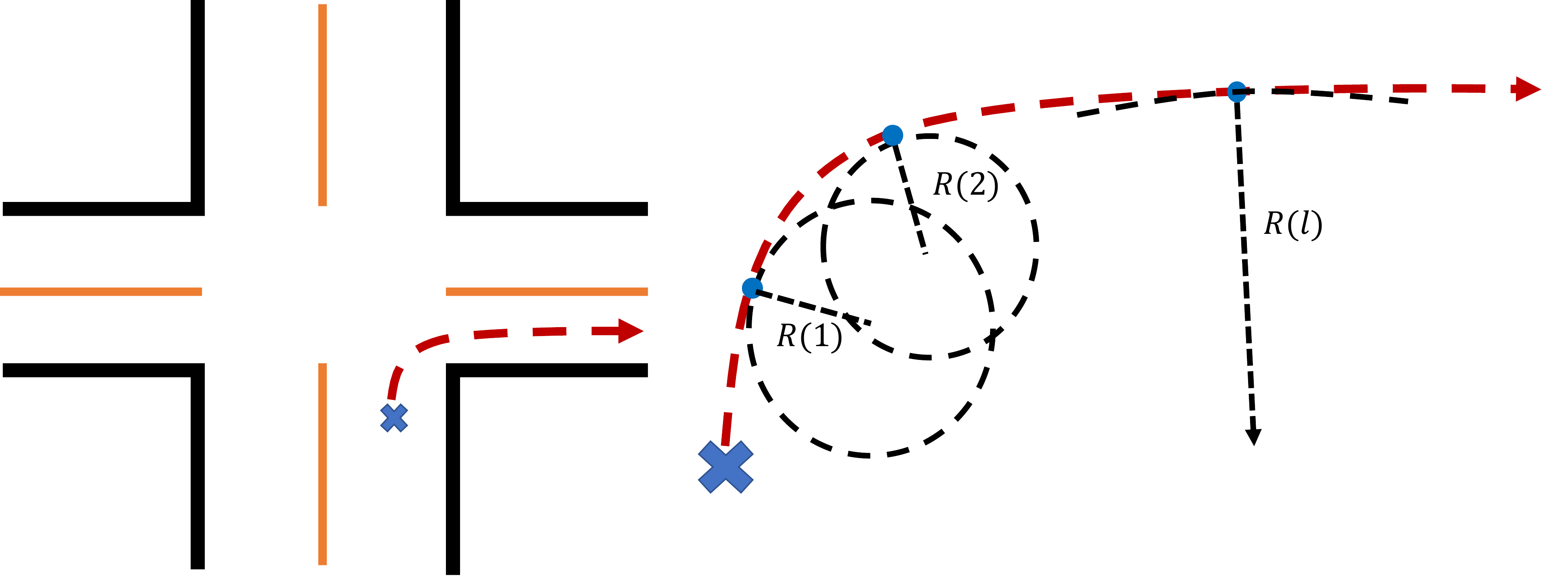}
\end{center}
   \caption{Illustration of the planned path description. Points are sampled (per distance unit) on the planned path obtained from the AD path planning module. Radius of curvature can be used to describe each point. Thus, a path can be represented by a discrete set of point descriptions.}
   \label{fig: goal}
\end{figure}

\textbf{Goal-oriented Feature}. At each time step, the planned path (with regard to distance in the vehicle-centric coordinates) can be obtained from the AD path planning module for an online driving task. As shown in Fig.~\ref{fig: goal}, at each time step, discrete points are uniformly sampled with respect to distance to represent the planned path. Each sampled point is represented by $(x, y)$ which indicates the location of the point in the vehicle-centric coordinate in the real world. Radius of curvature, $R$, is directly related to the turning behavior, so it can be used to represent each point on the path which can be calculated as in Eq.~\ref{eq: R_gps} given the location $(x,y)$. For the straight road, the value of $R$ approaches infinity which is not appropriate for learning. So, we use $IR=\frac{1}{R}$ instead to describe a certain point in the planned path. At time $t$, $\textbf{IR}^t = [IR(1), IR(2),...,IR(L)]$ is used to represent the whole planned path where $IR(l)$ indicate the value of $IR$ at the next $l$ distance units and $L$ indicates the maximum future distance our method considers. One FC layer is applied on $\textbf{IR}^t$ to extract the goal-oriented feature, $\textbf{g}^t$.
\begin{equation}
\label{eq: R_gps}
    R = sign\times(\frac{(1+y^{'2})^{\frac{3}{2}}}{y^{''}}),\\
\end{equation}
where $y^{'}=\frac{dy}{dx}$ and $y^{''}=\frac{d^2y}{d^2x}$. $sign=1$ when turning right and $sign=-1$ when turning left.
\section{Experiments}
\label{sec: Exp}
\subsection{Object Importance Estimation Dataset}
\textbf{Dataset Description}. We collect 743 on-road driving videos at traffic intersections in the real world. Data collection was conducted from two different locations- Mountain View and Sunnyvale, CA, USA, totalling 6.3 hours. Each location contains 3 sessions of data. We believe that intersections contain more complicated driving scenarios and are more challenging for our task, so from each of the raw videos, a short video is trimmed. Each short video contains one pass of an intersection (25 meters before and after the intersection). After trimming, 2.7 hours of useful data are obtained. All the annotations and our experiments are conducted on the trimmed videos.

\textbf{Annotations}. When preparing the important object annotations, an annotator was asked to watch the on-road driving video and imagine he/she was driving the ego vehicle. All the objects that are relevant for the ego vehicle's control decision are tightly located using bounding boxes. Note that the annotator was given the driving goal during the process of annotating each video sequence. For each video, important objects are labeled at every 30 frames. The frame sampling rate is 30 fps, thus labels were acquired at every second.

Further more, in order to understand our performance on different driving goals, \emph{i.e.}, \emph{turn left}, \emph{straight pass} and \emph{turn right}, per-frame goal are annotated. The goal of an image frame is annotated as `turn left' if the vehicle is expected to turn left at the next frame and so on.  

\textbf{Dataset Preprocessing}. Important object labeling may be influenced by traffic signals. For example, when the red light is on, no objects are considered as important since none of them will influence the driver's control decision. However, since we only consider the interactions with road users, we remove all the image frames where no important objects are labeled because of the traffic signals. 

\textbf{Dataset Statistics}. After preprocessing, $8,166$ image frames are annotated, where $4,268$ important objects are obtained. Among all the labeled frames, $56.6\%$ images contain no important objects, $38.3\%$ contain one important object and $5.1\%$ frames include multiple important objects.

The annotated frame numbers of \emph{turn left}, \emph{straight pass} and \emph{turn right} are $1004$, $6591$ and $1016$. The corresponding object numbers are $375$, $3573$ and $320$. 
Although we focus on traffic intersections, there are still more straight-pass frames than left/right-turn ones, which motivates us to evaluate the models based on different goals in order to avoid the results being dominated by the straight-pass scenario.

\begin{figure}
\begin{center}
   \includegraphics[width=0.9\linewidth]{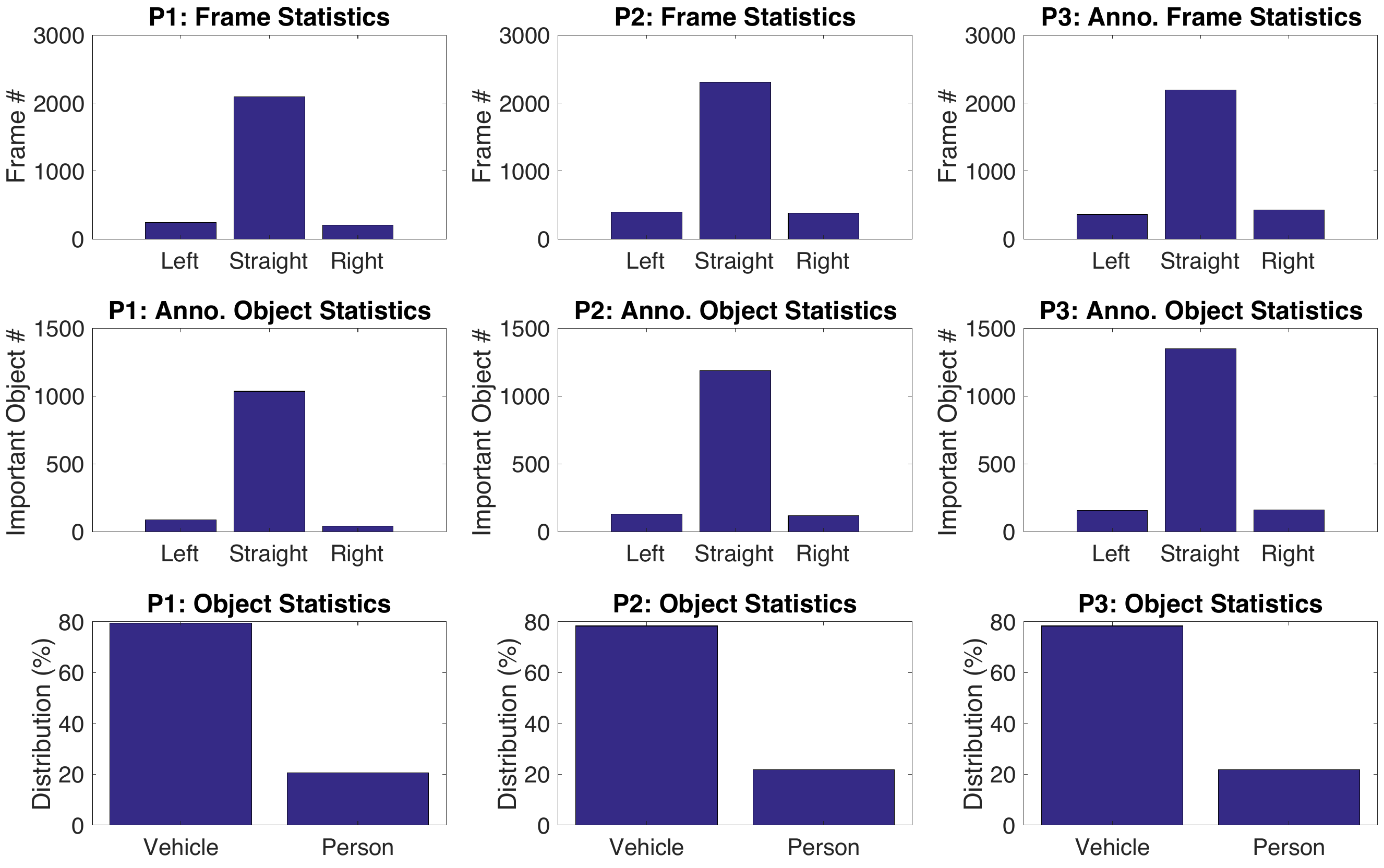}
\end{center}
   \caption[Caption for LOF]{Statistics of the split parts. The 1st and 2nd rows show the annotated frame and important object numbers based on different per-frame goals. The 3rd row shows the number percentages of vehicles and persons.}
\label{fig: anno_splits}
\end{figure}

\textbf{Train/test sets and statistics}. The dataset with 6 sessions is grouped into three parts~\footnote{We use 3-fold cross validation instead of 10-fold due to not enough data.}, \emph{i.e.}, P1, P2 and P3. For cross validation, all models are evaluated at every part while trained on the other two parts. We ensure that data of each part was collected from different sessions, locations and times, and has similar amount of videos and category distributions of road users~\footnote{Since we do not have the object-category annotations. We use the result of object detection (with confidence threshold of $0.5$) to estimate the numbers of vehicles and persons at the annotated frames.}. Tab.~\ref{tab: dataset} and Fig.~\ref{fig: anno_splits} show characteristics of each part. As shown, different parts have very similar statistics. 
\subsection{Planned Path Approximation}
\label{sec: pathEstimate}
Since the experiments are done in an off-line manner, data from the AD path planning module is not available. To evaluate our method, we recover (approximate) the planned path of our vehicle at a given time step as $\textbf{IR}^t\approx\hat{\textbf{IR}}^t=[\hat{IR}(1),\hat{IR}(2),...,\hat{IR}(L)]$ where $\hat{IR}(l)$ is calculated as in Eq.~\ref{eq: R}. We believe that it is easy to replace $\hat{\textbf{IR}}$ with $\textbf{IR}$ when AD path planning module is available.
\begin{equation}
\label{eq: R}
    \hat{IR}(l) = \frac{\omega(l)}{v(l)}
          = \frac{\alpha \times yr(l)}{v(l)},
\end{equation}
where $\omega(l)$, $v(l)$ and $yr(l)$ indicates angular velocity, velocity (kilometers per hour) and yaw rate (angle per second) at the next $l$ distance unit. One distance unite is $\frac{1}{3.6}$ meters. $\alpha$ is a scale number.

Both yaw rate and velocity can be obtained from the CAN bus sensors. Yaw rate values are negative when turning left while positive when turning right. 

\begin{figure}
\begin{center}
   \includegraphics[width=0.9\linewidth]{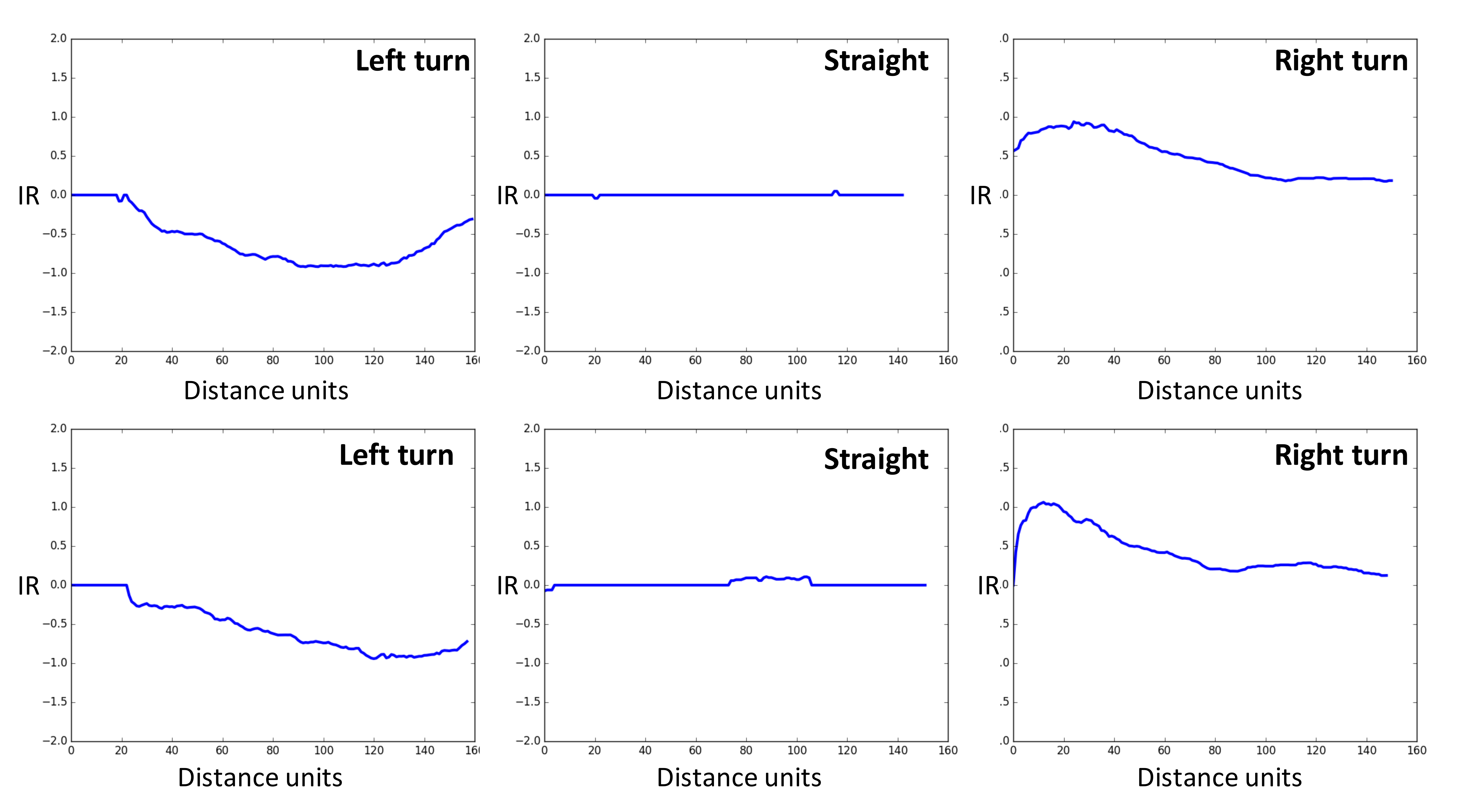}
\end{center}
   \caption{Examples of the $\hat{\textbf{IR}}$ given different driving goals.}
\label{fig: r}
\end{figure}

Examples of $\hat{IR}(l)$ for left turn, straight and right turn are shown in Fig.~\ref{fig: r}. As we can see, there are obviously discriminative patterns among the three driving goals, \emph{e.g.}, left turns have negative troughs, right turns have positive crests and straights are around zero .
\begin{table}[]
\setlength{\tabcolsep}{5pt}
\centering
\caption{Overall statistics of the split parts (P1, P2  and P3).}
\begin{tabular}{|c|cc|cc|cc|c|}
 & \multicolumn{2}{|c|}{P1} & \multicolumn{2}{|c|}{P2} & \multicolumn{2}{|c|}{P3}& Total \\
Session & S1 & S2 & S3 & S4 & S5 & S6 & NA\\
Location & MV & SV & MV & SV & MV & SV & NA\\
Video \# & 134 & 100 & 183 & 87 & 188 & 51 & 743 \\
\multicolumn{1}{|c|}{Anno. Frame \#} & \multicolumn{2}{|c|}{2,541} & \multicolumn{2}{|c|}{3,087} & \multicolumn{2}{|c|}{2,983}& 8,611\\
\multicolumn{1}{|c|}{Anno. Obj \#} & 
\multicolumn{2}{|c|}{1,164} & \multicolumn{2}{|c|}{1,436} & \multicolumn{2}{|c|}{1,668} & 4,268
\end{tabular}
\label{tab: dataset}
\end{table}
\begin{table*}
\setlength{\tabcolsep}{4pt}
\centering

\caption{Comparison between our Goal-Visual model and the baselines in terms of Average Precision (\%) on \emph{turn left} (Lt), \emph{straight pass} (St), \emph{turn right} (Rt) and all (All) frames. Avg. indicates the average value of the corresponding results among P1, P2 and P3.}
\begin{tabular}{|c|cccc|cccc|cccc|cccc|cccc|}
 & \multicolumn{4}{|c|}{P1} & \multicolumn{4}{|c|}{P2} & \multicolumn{4}{|c|}{P3} & \multicolumn{4}{|c|}{Avg.} \\
Test Set & Lt & St & Rt & All & Lt & St & Rt & All & Lt & St & Rt & All & Lt & St & Rt & All \\
\hline
Visual Model-Image & 23.5 & 42.9 & 16.1 & 35.5 & 22.9 & 42.7 & 26.7 & 42.1 & 19.1 & 33.9 & 25.3 & 32.6& 21.8 & 39.8 & 22.7& 36.7 \\
Visual Model & 35.8 & 71.2 & 34.7 & 68.1 & 56.0 & 70.6 & 54.2 & 68.1 & 36.4 & 72.4 & 57.4 & 70.9 & 42.7 & 71.4 & 48.8 & 69.0 \\
Goal-Geometry Model & 41.1 & 32.9 & 22.8 & 32.1 & 32.5 & 42.6 & 19.7 & 40.6 & 25.6 & 45.8 & 30.2 & 41.8& 33.1 & 40.4 & 24.2& 38.2 \\
\textbf{Goal-Visual Model} & \textbf{48.9} & \textbf{72.2} & \textbf{42.8} & \textbf{70.2} & \textbf{61.1} & \textbf{71.7} & \textbf{70.3} & \textbf{70.3} & \textbf{45.2} & \textbf{75.8} & \textbf{61.7} & \textbf{72.0} & \textbf{51.7} & \textbf{73.2} & \textbf{58.3} & \textbf{70.8} \\
\hline
\hline
Random Chance & 4.3 & 5.2 & 2.7 & 4.8 & 5.3 & 5.9 & 14.0 & 8.4 & 5.7 & 6.7 & 4.7 & 6.1 & 5.1 & 5.9 & 7.1 & 6.4 \\
UpperBound & 90.9 & 81.6 & 72.7 & 81.7 & 90.9 & 81.7 & 90.8 & 90.8 & 90.4 & 89.2 & 90.9 & 89.4 & 90.7 & 84.1 & 84.8 & 87.3
\end{tabular}
\label{tab: quantitative}
\end{table*}

\subsection{Baselines}
\textbf{Upperbound}. We estimate importance scores for all the object proposals (tracklinks), so the final results depend on the quality of the detection and tracking algorithm. We assign the correct importance label for each proposal link in this baseline. Thus, it is the upper bound of our method and all the mistakes are due to the bad detection and tracking. 

\textbf{Random Chance}. We randomly assign a value ($\in [0,1]$) to each proposed tracklink as its important probability in this baseline. So, it is the lower bound of our method.

\textbf{Visual model}. It contains only the first branch of our framework which has only the visual features as input to the LSTM model. We want to see how the goal information can improve the prediction results quantitatively.

\textbf{Visual model-Image}. This model does not utilize the temporal information and predicts object importance scores by just observing the target image frame. In order to do that, we replace the LSTM model with one FC layer. This baseline is to compare with the standard object detection framework and evaluate how much the temporal information can help.

\textbf{Goal-Geometry Model}. This baseline has the same two-branch structure as our method except that appearance feature is removed and only motion and location features are used. Comparing it with our method will show if the method performs good if semantic local context is not given.

\subsection{Implementation Details}
\label{sec: implementation}
Tracking-by-detection~\cite{andriluka2008people} framework is used to conduct object tracking, where Faster R-CNN~\cite{ren2015faster} with Resnet101 is used for detection and SORT~\cite{bewley2016simple} is used for tracking. Some of the objects may not start at the first frame or last till the end. We only keep the objects that still exist at the last frame and pad $0$s in the front if they do not start at the first frame.

The length of video clip, $n$, is set to $30$. We set $L$=40 which is roughly 10 meters in the real world. $\alpha$ in Eq.~\ref{eq: R} is set to 1. For the visual model, we set length of the LSTM hidden layer to be $256$ and the FC layer in goal model is set to be $16$. For image based visual model, the FC layer has $1,024$ units. Weighted-cross-entropy loss is used to optimize our model and all the baselines. The weights for positive and negative samples are inversely proportional to their sample numbers in one training batch.

\subsection{Experimental Results}
\label{sec: quantitative}
Comparisons between our method, \emph{i.e.}, \emph{Goal-Visual Model}, and the baselines using \emph{average precision (AP)} are shown in Tab.~\ref{tab: quantitative}. Our method largely outperforms \emph{Random Chance} (``by-chance" approach). Comparing \emph{Visual Model} with \emph{Visual Model-Image}, we see that the temporal information is essential for our task. Without temporal modelling, the overall \emph{AP} drops by $32.3\%$. With the goal information, our \emph{Goal-Visual Model} outperforms the \emph{Visual Model} by about $2\%$ in terms of \emph{AP}.
\begin{table*}
\centering
\setlength{\tabcolsep}{4pt}
\caption{Comparison between our model and the baselines in terms of mean Average Precision (\%) based on different object categories. Pn and Ve means `person' and `vehicle'. Avg. indicates the average value of the corresponding results among P1, P2 and P3.}
\begin{tabular}{|c|ccc|ccc|ccc|ccc|}
 & \multicolumn{3}{|c|}{P1} & \multicolumn{3}{|c|}{P2} & \multicolumn{3}{|c|}{P3} & \multicolumn{3}{|c|}{Avg.} \\
 & Pn & Ve &mAP & Pn & Ve&mAP & Pn & Ve&mAP & Pn & Ve&mAP \\
 \hline
Visual Model-Image & 17.7 & 42.6 & 30.15 & 29.6 & 46.9 & 38.25& 21.2 & 39.8 & 30.5&22.8 & 43.1&33.0 \\
Goal-Geometry Model & 34.8 & 35.3&35.1 & 36.9 & 44.4& 40.7& 45.2 & 43.6& 44.4& 40.0 & 41.1&40.6 \\
Visual Model & 56.0 & 75.4&65.7 & 56.1 & 76.4 &66.3& 49.7 & \textbf{78.6} &64.2& 53.9 & 76.8&65.4 \\
\textbf{Goal-Visual Model} & \textbf{60.0} & \textbf{76.2}& \textbf{68.1}& \textbf{61.2} & \textbf{78.1}& \textbf{69.7}& \textbf{57.6} & 77.3 &\textbf{67.5}& \textbf{59.6} & \textbf{77.2}&\textbf{68.4}
\end{tabular}
\label{tab: cate}
\end{table*}

To evaluate the effectiveness of local visual scene context, our method is compared with \emph{Goal-Geometry Model}. The \emph{Goal-Geometry Model} only captures the motion and location information of a road user and combines it with the goal of the ego vehicle, without knowing the scene semantic. As it is shown, our method largely outperforms this baseline which demonstrates the usefulness of the scene context.

To evaluate our performance on different driving goals, we validate our method and the baselines on \emph{turn left}, \emph{straight pass} and \emph{turn right} frames separately. Intuitively, our goal model should help more on the \emph{turn left} and \emph{turn right} cases compared to the \emph{straight pass}. From the results in Tab.~\ref{tab: quantitative}, our method largely improves the \emph{Visual Model} by $9\%$ \emph{AP} for \emph{turn left} and by $9.5\%$ for \emph{turn right}.
\begin{figure*}
\begin{center}
   \includegraphics[width=0.9\linewidth]{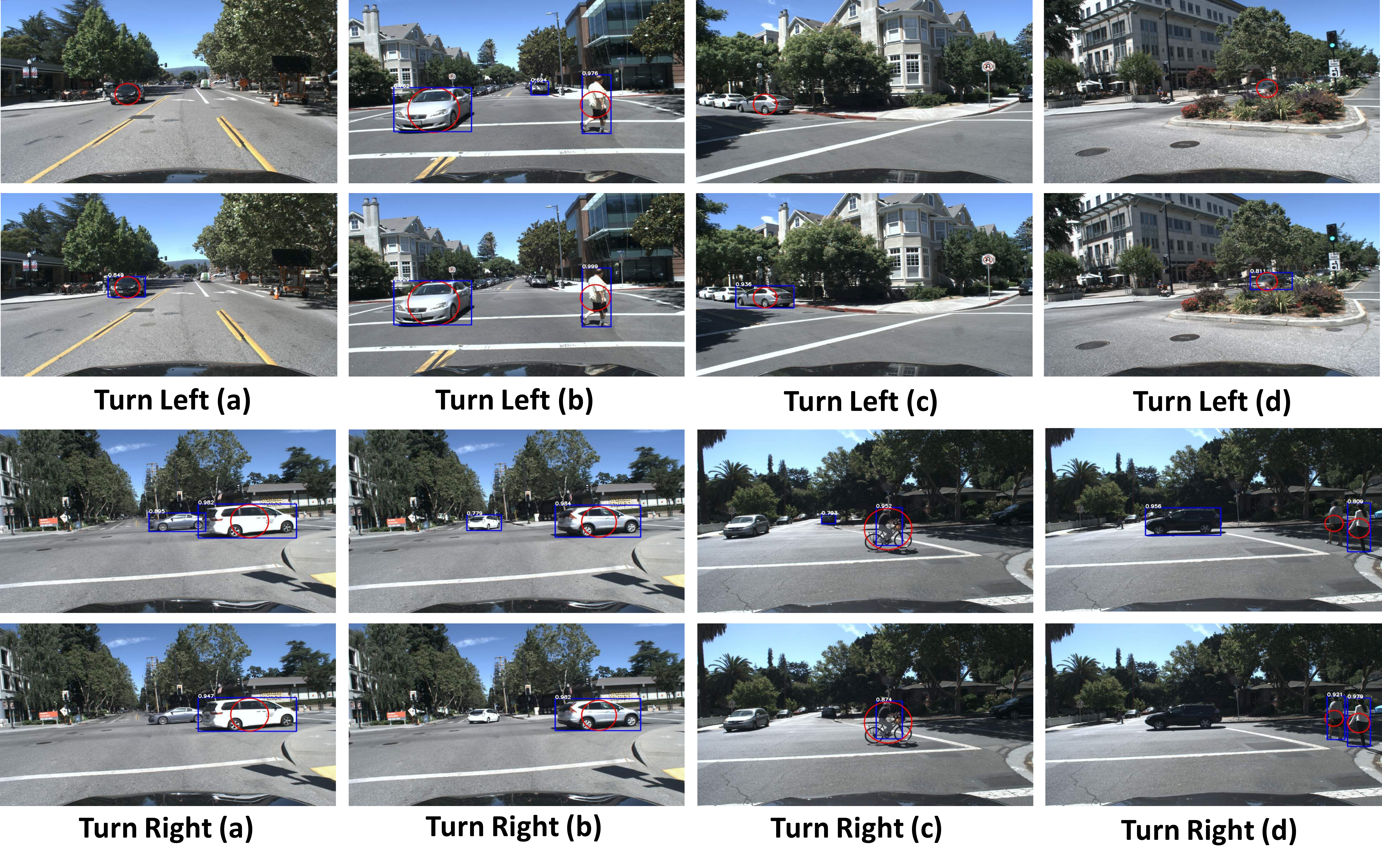}
\end{center}
 \vspace{-1em}
   \caption{Qualitative comparisons between the \emph{Visual Model} ($1st$, and $3rd$ rows) and our \emph{Goal-Visual Model} ($2nd$ and $4th$ rows) on the \emph{turn left} and \emph{turn right} frames. The red circles indicate ground truth, the blue boxes are the detected objects with the importance scores. For concise visualization, only objects with more than 0.5 importance scores are shown.}
\label{fig: qualitative}
\end{figure*}

We are also interested in our performance on different object categories, \emph{i.e.}, person and vehicle. Since, we do not have ground truth of the object categories, we generate the class label using the detection results. We match each labeled important object to a detected object if they have the largest Intersection over Union (IoU) and the $IoU>0.5$. It is not guaranteed that every important object will find a match, since the detector is not perfect. However, experiment shows that around $95\%$ of important objects are matched, so we ignore the small amount of unmatched ones. Comparisons between our method and the baselines are shown in Tab.~\ref{tab: cate}, which demonstrates that our model outperforms all the baselines in terms of \emph{mAP}. Specifically, we observe that performance on the `person' category is largely improved with goal information. \emph{Goal-Visual Model} improves by around $6\%$ on `person' compared to \emph{Visual Model}. It may due to the fact that most important persons are those who are walking cross the road. It is essential for the model to know where the ego vehicle is going in order to infer if a pedestrian on a certain side is important. 

Qualitative results on \emph{turn left} and \emph{turn right} are shown in Fig.~\ref{fig: qualitative}. As it is shown, knowing the driving goal can help capture important objects on (or coming to) our future path, \emph{e.g.}, \emph{turn left(a)(c)(d)} and  \emph{turn right(d)}. It can also filter out objects that are impossible to block our way based on their motion and location, \emph{e.g.},  \emph{turn left(b)} and \emph{turn right(a)(b)(c)}.

\begin{figure}
\begin{center}
   \includegraphics[width=0.9\linewidth]{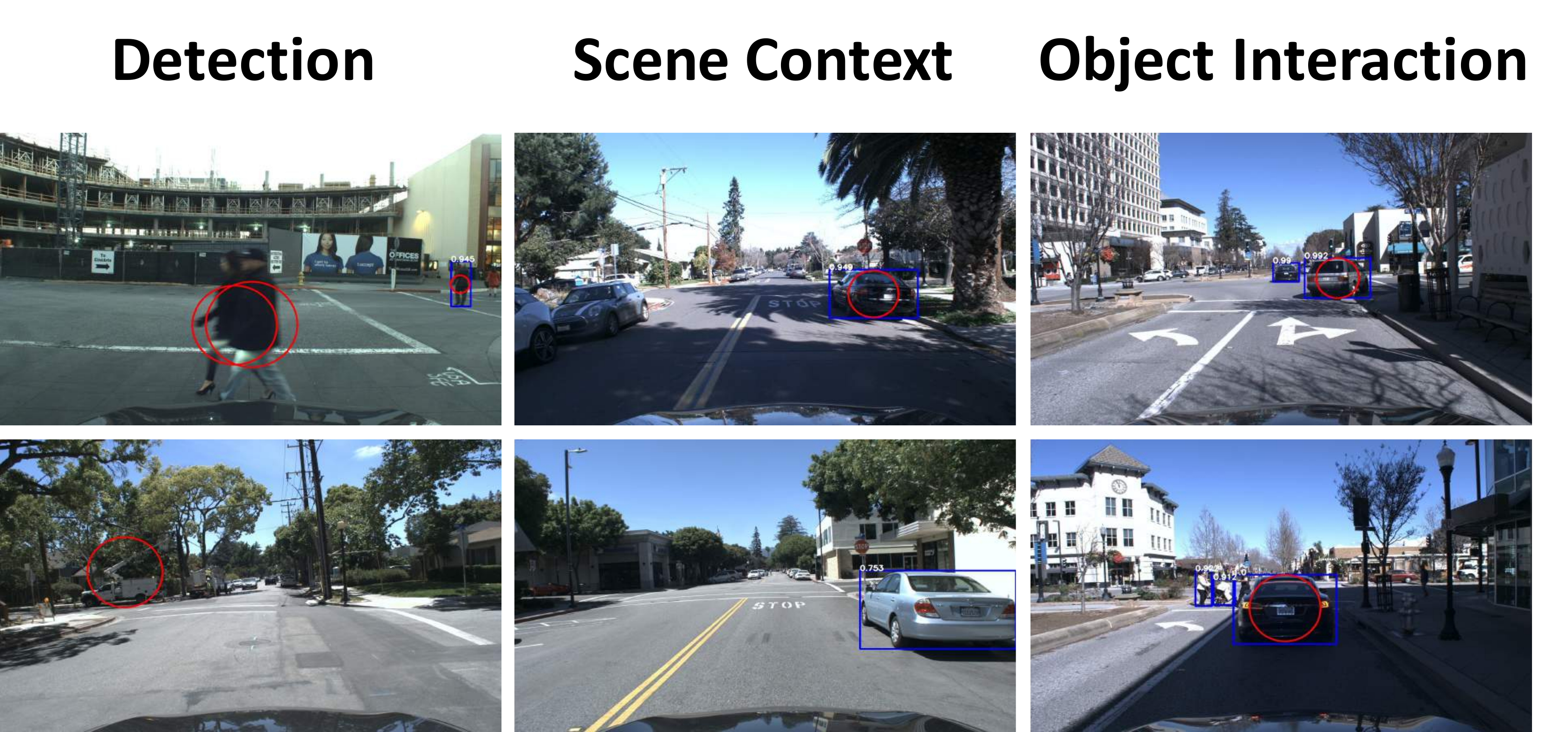}
\end{center}
   \caption{Major failure cases of our method. The examples of the $1st$ column are due to miss detection, those of the $2nd$ column is due to the lack of global scene context and the $3rd$-column ones are because of the lack of the interaction among road users. Red circle and blue box indicate ground truth and our result, respectively.}
\label{fig: failure}
\end{figure}

Three major failure cases are shown in Fig.~\ref{fig: failure}. The first one is because of the bad detection/tracking results. When the detection of the important object fails, there is no way for our framework to correct it. That is why our upper bound is not $100\%$ \emph{AP}. The second case is a result of missing global scene context. The comparison shows that for the two parked car, one is thought as important, but the other one is not. Based on our observation, the annotator tends to annotate the parked car if the road is narrow. The third case is due to the lack of communication among road users. For example, if we remove the labeled car in the last image, all the pedestrians should be important. They are not labeled as important because there is a closer car stopping the ego vehicle hitting them. Since our method does not model the interactions among road users, it is hard for an object to know the status of other objects. Future works are needed to solve these three failure cases.

\subsection{Are Road Users Equally Important?}
For a proof-of-concept, we propose a binary brake prediction (BBP) framework with object importance as a input.

BBP is a simplified version of brake prediction task which has binary labels, $y_{brake}$, instead of continuous brake values (can be obtained from CAN bus data), $v_{brake}$ ($y_{brake}=1$ if $v_{brake} > 0$ and $y_{brake}=0$ otherwise). The input of BBP is a video clip and output is the brake probability of the ego vehicle in the last frame.

We assume that brakes depend only on the interaction between the road users and the ego vehicle, since we have removed the traffic-light related frames from our dataset. The visual model in Fig~\ref{fig: pipeline} is used to predict brake score, $\textit{s}^t_i$, at time $t$ of the ego vehicle given road user,$i$, in the input video clip. The final brake score, $\textit{s}^t_{fuse} = \underset{i}{\sum}{(w^t_i* \textit{s}^t_i)}$, is obtained by fusing predicted scores based on all the road users in a weighted sum manner. Our model use the predicted important probability to be the weight of each object. Our intuition is that more important objects will have bigger impacts on the brake decision. The baseline uses the same weight ($0.5$) for all the objects to indicate that all objects in the scene equally contributed to the brake. 

Experimental results suggest that our method improves the baseline by $4.3\%$, $1.7\%$ and $1.3\%$ AP in the P1, P2 and P3, respectively, which demonstrates the potential usefulness of the object importance.

\section{Conclusion}
We propose a new problem as Object Importance Estimation (OIE) in on-road driving videos to understand the human visual selection mechanism under the driving context. We present a novel framework to handle the problem where both the visual dynamics of road users and the goal of the ego vehicle are taken into consideration. To evaluate the problem, we collect an on-road driving dataset and annotate the important objects given the video clip. Experimental results demonstrate the effectiveness of our idea. Moreover, we explore the potential usage of the OIE by incorporating it into a binary brake prediction framework. Experiments show that important objects can help to improve the prediction.
{\small
\bibliographystyle{ieee}
\bibliography{egbib}
}
\end{document}